\documentclass[letterpaper]{article} 

\usepackage{aaai24}  
\usepackage{times}  
\usepackage{helvet}  
\usepackage{courier}  
\usepackage[hyphens]{url}  
\usepackage{graphicx} 
\usepackage{csquotes}
\usepackage{natbib}  
\usepackage{caption} 
\usepackage{bibunits}
\usepackage{algorithm}
\usepackage{algorithmic}
\usepackage{multirow}
\usepackage{newfloat}
\usepackage{listings}
\DeclareCaptionStyle{ruled}{labelfont=normalfont,labelsep=colon,strut=off} 
\floatstyle{ruled}
\newfloat{listing}{tb}{lst}{}
\floatname{listing}{Listing}
\title{Exploring Nature: Datasets and Models for Analyzing Nature-Related Disclosures}
\author{
    Tobias Schimanski\textsuperscript{\rm 1}, Chiara Colesanti Senni\textsuperscript{\rm 1}, Glen Gostlow\textsuperscript{\rm 1}, \\
    Jingwei Ni\textsuperscript{\rm 1, \rm 2}, Tingyu Yu\textsuperscript{\rm 1}, Markus Leippold\textsuperscript{\rm 1,3}\\
}
\affiliations{
    \textsuperscript{\rm 1}University of Zurich  \quad 
    \textsuperscript{\rm 2}ETH Zurich \quad 
    \textsuperscript{\rm 3} Swiss Finance Institute (SFI)
}

\usepackage{bibentry}

\begin{document}

\maketitle

\begin{abstract}
Nature is an amorphous concept. Yet, it is essential for the planet's well-being to understand how the economy interacts with it. To address the growing demand for information on corporate nature disclosure, we provide datasets and classifiers to detect nature communication by companies. We ground our approach in the guidelines of the Taskforce on Nature-related Financial Disclosures (TNFD). Particularly, we focus on the specific dimensions of water, forest, and biodiversity. For each dimension, we create an expert-annotated dataset with 2,200 text samples and train classifier models. Furthermore, we show that nature communication is more prevalent in hotspot areas and directly effected industries like agriculture and utilities. Our approach is the first to respond to calls to assess corporate nature communication on a large scale.
\end{abstract}

\section{Introduction}
\textbf{Motivation.}
It is well known that modern economies are threatened by climate change and that there are feedback effects between climate change and the economy. However, much less is known with regard to the economic impact of other nature-related threats, such as water stress and water pollution, deforestation, biodiversity loss, invasive species, and soil degradation. Similarly, the negative impact of economic activities on nature loss has only been marginally explored \citep{ngfs2022biosphere}. Studying the relationship between nature and the economy is complex for several reasons. First, nature involves several dimensions and cannot be reduced to a single metric such as CO$_2$ emissions. Second, the consequences of nature loss tend to be local and context-specific. Third, climate change and nature loss are deeply interconnected and mutually reinforcing. For example, deforestation is not only harmful to biodiversity but also increases climate risks. Companies' disclosure provides insights into their perception of risks and opportunities, as they are in the best position to assess them. Modern Natural Language Processing (NLP) can help to simplify and structure analyses of companies' disclosures. Using this information can improve our understanding of the interaction between nature and the financial system as well as the broader economy.

\textbf{Contribution.}
To better understand these interactions, this paper delivers three contributions. First, we create a 2,200 text samples dataset for the detection of communication in the nature dimensions of water, forest, and biodiversity. We further form a general nature dimension combining the three subdimensions\footnote{We are aware of the fact that nature includes many other dimensions beyond the ones considered in our approach. However, for simplicity, we adopt this terminology here.}. Second, we develop machine learning models to detect language patterns in company disclosures. We fine-tune various transformer-based architectures and compare their performance. Third, we propose a use case to analyze the current nature communication of companies. In this use case, we assess the 2021 cross-section of earning calls and investigate country- and industry-specific differences.\footnote{Both models and datasets are available on \url{https://huggingface.co/ESGBERT} .}

\textbf{Results.}
Using the 2,200 text samples dataset for each nature dimension, we show that further pre-trained BERT models on the climate- and environmental domain slightly outperform their counterparts pre-trained on general domain language. Furthermore, countries and industries in nature hotspots tend to contain companies that discuss these topics more, validating our approach.

\textbf{Implications.}
Our findings yield significant implications for both academics and professionals. With a rapid increase in the volume of disclosed textual information, there is a growing demand for precise and actionable metrics aligned with nature principles. The datasets and models developed within this paper can help investment professionals, analysts, and strategists analyze companies' behavior. This delivers a valuable complement to existing third-party data. Furthermore, knowing that regulatory control mechanisms are in place might incentivize companies to deliver actions on their communication. Finally, these datasets and models aim to assist the academic community in future studies that try to understand and explain the interaction between nature and the economy.

\section{Background}
Climate change and nature loss, like deforestation, water pollution, and biodiversity loss, are increasingly recognized as threats to the global economy, while awareness about the negative impacts of economic activities on nature is rising. The concept of double materiality – which captures the interactions between the financial system and ecosystems – is particularly important in understanding how economies should react to the growing challenges associated with the environmental crisis. 

\textbf{Rising Importance of Nature in Finance.}
Both the practice and research community lay an increasing focus on nature-related topics, especially in the realm of finance. For instance, the UN recently started discussions to introduce biodiversity credits and HSBC announced a nature fund that raised over 650 million USD. Additionally, new initiatives like the Taskforce on Nature-related Financial Disclosures (TNFD) have been formed to guide company disclosure on nature.

From a research perspective, accounting for nature-related risks enables financial markets to achieve a more efficient allocation of resources. However, the concept of nature in the accounting and finance literature is poorly understood, yet growing in importance \cite{ngfs2022biosphere, karolyi2023}. Some recent empirical contributions have shown that financial markets only recently price biodiversity \cite{garel2023, coqueret2023} and misprice water risks \citep{colesanti2023economic}. Moreover, managing biodiversity risk appears to lower refinancing costs \cite{hoepner2023}. However, these studies mainly rely on data providers and geospatial information to assess nature.

\textbf{NLP in the Nature Domain.}
Disclosure provides an important source of nature-related information for financial markets. NLP offers a tool to extract and organize such information efficiently. An infant finance literature exists that analyses nature risks using textual disclosures. For instance, \citet{giglio2023} rely on a keyword-based approach. Keyword approaches are also common in the general climate-related finance literature \cite[e.g.,][]{sautner2023, nagar2022}. However, prior research has already outlined that these methods suffer from a lack of context-sensitivity \cite{varini2020}. For example, the term \enquote{nature} itself can include alternating meanings depending on the context.

Recent work has acknowledged these shortcomings and instead utilizes machine learning architectures like transformer-based BERT models. These models profit from learning language patterns based on a large number of text samples and gain a semantic understanding of texts \citep{devlin2019bert}. Prior research has pre-trained BERT models on climate- and environmental texts to enhance the models' language understanding for these specific domains \citep{webersinke2021, schimanski2023bridiging}, as well as fine-tuned models for downstream tasks. For instance, BERT models have been used to detect climate-related patterns in text \cite{stammbach2022dataset, luccioni2020}, general environmental, social, and governance (ESG) discussions \cite{schimanski2023bridiging}, as well as cheap talk and potential greenwashing patterns \cite{bingler2022}. 

There is also a growing trend leveraging Large Language Models (LLMs) to parse sustainability-related documents \citep{ni2023chatreport,vaghefi2023chatclimate}. But LLMs usually suffer from the high cost of hardware usage, making them a suboptimal choice for analyzing large amounts of text compared to lighter LM classifiers. However, none of the previous work has set a focus on creating a comprehensive machine-learning approach to assess nature. 

\textbf{Nature-Related Disclosures and Regulation.}
In line with a steady rise in the perceived importance of nature, there exists an increasing amount of disclosure frameworks and obligations for companies. One of the most important frameworks is the Corporate Sustainability Reporting Directive (CSRD) of the European Union. Introduced at the beginning of 2023, this directive strengthens the rules concerning the environmental information that companies have to report. From 2024 on, companies are also mandated to disclose in line with the European Sustainability Reporting Standards (ESRS). The standards cover the full range of environmental, social, and governance domains, including climate change, biodiversity, and human rights. They provide information for investors to understand the sustainability impact of the companies in which they invest. 

Furthermore, a more global approach is taken by the Taskforce on Nature-related Financial Disclosures (TNFD). The initiative has developed a set of disclosure recommendations and guidance for organizations to report and act on evolving nature-related dependencies, impacts, risks, and opportunities. The recommendations aim to enable companies and investors to integrate nature into decision-making and ultimately support a shift in global financial flows away from nature-negative outcomes. In this paper, we orient on the TNFD guidelines and make use of their nature dimensions water, forest, and biodiversity. 


\section{Data Creation}
To create classifiers for the nature domain, we compile a 2,200-sentence dataset in which the texts are assigned to either one or multiple of the dimensions water, forest, or biodiversity. The major problem arising in the creation of such a dataset represents the scarcity of nature-related text samples in company disclosures. While the topic's importance rises steadily, it still represents a small minority class. The final dataset needs to represent both a sufficient amount of true nature-related sentences as well as edge cases and non-nature-related sentences for the models to obtain an accurate understanding of the decision boundary between the classes. To address this problem, we create the final dataset in five steps. 

First, we create a general base dataset comprising annual reports, sustainability reports, and earning calls transcripts. We chose to include both written (reports) and verbal (transcripts) communication in our data creation process to consider the bandwidth of both communication forms. Collectively, we gather over 25 million sentences for our base dataset (see Appendix A\ref{appendix:datacreation_steps_overview} for more details). In the second step, we filter a subset of the base data with keywords related to each of the nature dimensions water, forest, and biodiversity (see Appendix B\ref{appendix:keywords_sampeling} for more details). We intentionally use a broad spectrum of keywords to account for both actual cases, edge cases, and generic environmental cases that should not be considered as nature-related. Therefore, the keyword-filtered samples may contain a large amount of false positive cases that just contain a specific keyword. To overcome this issue, in the fourth step, we use GPT-3.5 to pre-label another subset of these filtered sentences. We use the labeling guidelines (see Appendix D) to design prompts for the Large Language Model. These prompts assign a score between 0-100 to each sentence depending on how closely the model perceives the text sample with the respective dimension. This allows a targeted sampling of sentences for each dimension. We both sample seemingly irrelevant and relevant sentences into the labeling dataset (see Appendix C\ref{appendix:chatgpt_prompt}). Finally, we construct a dataset with 2,200 text samples and label every sentence with four human expert annotators towards water, forest, and biodiversity. Split decisions in the labeling process are discussed and resolved within the annotator team. Furthermore, we calculate the inter-annotator agreements to gain an intuition about the task complexity. The agreement rates are consistently very high with Fleiss' Kappas of around 80\%. This signals that the task is easy to solve for the majority of the samples (see Appendix E\ref{appendix:annotator_agreement} for inter-annotator agreements).

\begin{figure}[h]
        \centering
	\includegraphics[width=.45\textwidth]{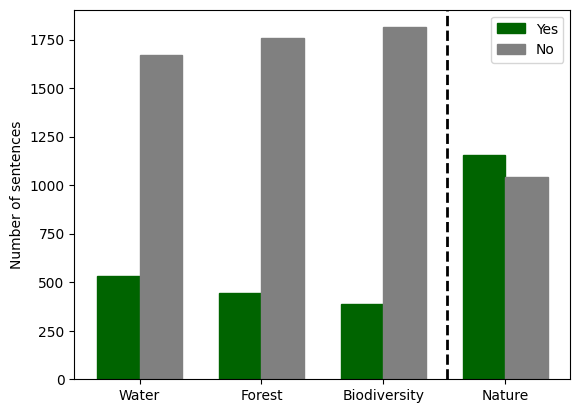}
 \caption{Distribution of the labeling data}
 \label{fig:naturelabeldist}
\end{figure}

Collectively, this process allows us to create an expert-annotated dataset addressing the three nature dimensions water, forest, and biodiversity. Furthermore, we argue that if any of these subdimensions of nature apply to a text sample, we can assign the general nature label to it. Figure \ref{fig:naturelabeldist} shows an overview of the label distribution in the dataset. Since nature sentences are scarce in the real world, all single nature dimensions are minority classes in the datasets.

\section{Nature Models}
To create the models, we fine-tune four pre-trained BERT models with our datasets. First, we use models pre-trained on general language corpus like RoBERTa \citep{liu2019roberta} and DistilRoBERTa \citep{sanh2020distilbert}. The main difference between these models is their size. While RoBERTa possesses 125 million parameters, DistilRoBERTa is a more resource-efficient version of the RoBERTa with 80 million parameters \citep{sanh2020distilbert}. Furthermore, we use both ClimateBERT and EnvironmentalBERT which are specifically pre-trained on climate- and environmental-related text corpus to improve their performance on related downstream tasks. Both models are fundamentally based on the DistilRoBERTa architecture \citep{webersinke2021, schimanski2023bridiging}. 

To evaluate the performance of the models, we use five-fold cross-validation (see Appendix F for the base hyperparameter setup). This allows us to validate the performance using the entirety of the data. Furthermore, we use the F1-score as a primary evaluation metric to assess the models. The F1-score is particularly useful for imbalanced datasets like ours because it takes into account positive cases while minimizing false positives and false negatives.

As Table \ref{tab:modelf1scores} shows, all models perform similarly well. However, the further pre-trained models ClimateBERT and EnvironmentalBERT marginally outperform their base and larger counterparts. These results hold true for other performance metrics like accuracy, precision, and recall (see Appendix F\ref{appendix:modelresults}), different sets of hyperparameters (see Appendix G\ref{appendix:hypergrid}), and significantly outperform keyword-based approaches (see Appendix H\ref{appendix:biodivkeywords}). It becomes apparent that for all tasks, the models can build a decision boundary very well. This is consistent with the inter-annotator agreements during the labeling process.

\begin{table}[]
\centering
\begin{tabular}{ccc}
\hline
\textbf{Domain}                        & \textbf{Model}    & \textbf{F1-Score (std.)} \\ \hline
\multirow{4}{*}{\textbf{Water}}        & EnvironmentalBERT & 0.9447 (0.0137)          \\
                                       & ClimateBERT       & \textbf{0.9510 (0.0113)}  \\
                                       & RoBERTa           & 0.9455 (0.0086)          \\
                                       & DistilRoBERTa     & 0.9498 (0.0116)          \\ \hline
\multirow{4}{*}{\textbf{Forest}}       & EnvironmentalBERT & \textbf{0.9537 (0.0092)} \\
                                       & ClimateBERT       & 0.9534 (0.0046)          \\
                                       & RoBERTa           & 0.9478 (0.0048)          \\
                                       & DistilRoBERTa     & 0.9529 (0.0065)          \\ \hline
\multirow{4}{*}{\textbf{Biodiversity}} & EnvironmentalBERT & \textbf{0.9276 (0.0191)} \\
                                       & ClimateBERT       & 0.9249 (0.0134)          \\
                                       & RoBERTa           & 0.9246 (0.0154)          \\
                                       & DistilRoBERTa     & 0.9229 (0.0203)          \\ \hline
\multirow{4}{*}{\textbf{Nature}}       & EnvironmentalBERT & \textbf{0.9419 (0.0081)} \\
                                       & ClimateBERT       & 0.9350 (0.0046)           \\
                                       & RoBERTa           & 0.9397 (0.0026)          \\
                                       & DistilRoBERTa     & 0.9355 (0.0072)          \\ \hline
\end{tabular}
\caption{F1-scores of the models for the nature tasks}
\label{tab:modelf1scores}
\end{table}

\section{Case Study}
In order to test the models in a real-world setting and provide evidence for their validity, we also propose a case study. We use the fine-tuned EnvironmentalBERT models and analyze earning call transcripts of companies worldwide retrieved from Refinitiv. Earning calls are quarterly scheduled events in which companies communicate their financial results and performance to investors, analysts, and the general public. Specifically, we use the transcripts of earning calls in the year 2021, split them into sentences, and label each sentence with our models. Furthermore, we divide the number of sentences mentioning nature topics by the number of all sentences. Finally, we build a yearly average across all earning calls for each company. Then, we investigate country- and industry-specific differences.

\begin{figure}[h]
        \centering
	\includegraphics[width=.5\textwidth]{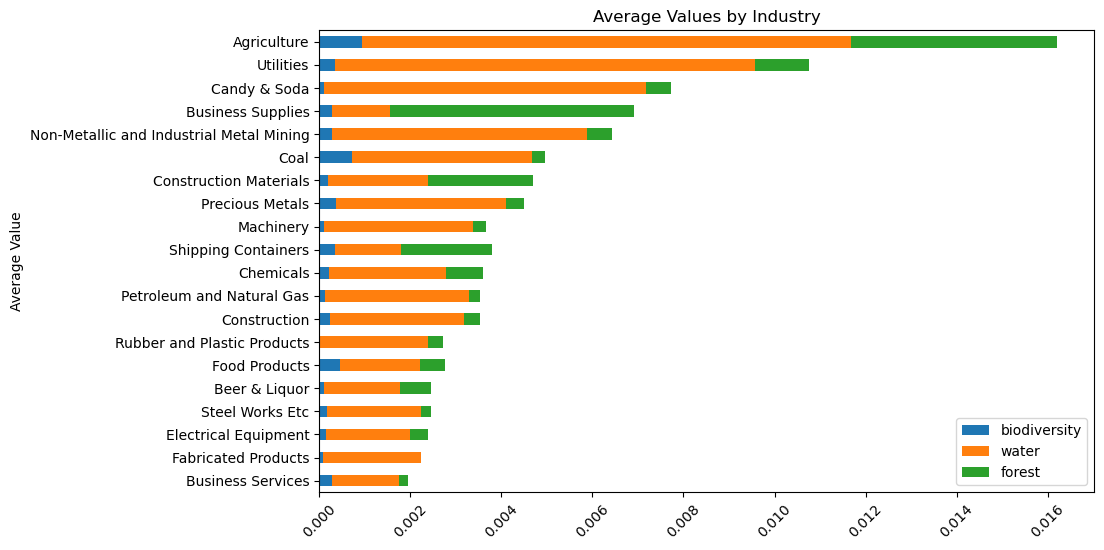}
 \caption{Top 20 industries communicating about nature-related topics measured as a ratio of nature communications vs. all communication}
 \label{fig:natureIndustries}
\end{figure}

As Table \ref{fig:natureIndustries} suggests, we find that resource-intensive industries such as agriculture and utilities have the highest nature exposure. While the water dimension seems to be the most important for the majority of industries, Business Suppliers and Construction Materials display a higher exposure to forest. To analyze country differences, we calculate the proportion of earning calls that mention nature-related topics at least once. Moreover, nature is mostly discussed in ecosystem hotspots near the equator (see Appendix I\ref{appendix:naturemention} for more details). While this exploratory exercise solidifies the validity of the method by showing that obvious patterns hold true, it only represents a first step and lays out the potential for future research.

\section{Conclusion}
This paper contributes to the understanding of the interaction between nature and the economy by developing new datasets and methods to perform a large-scale analysis of company-level nature communication. These can be used as a starting point for a variety of stakeholders to investigate companies' relationship with nature. Our tools can both ease the manual analysis of disclosures as well as automate steps towards assessing the real nature performance of companies.

\section{Limitations}
As with every work, this has limitations. While we provide an extensive evaluation that these models work on our dataset, the real-world data distribution is likely much more imbalanced. Extending the dataset with real-world (misclassified) samples for specific use cases will likely improve the performance. Furthermore, this could also solve a problem in the data creation. We pre-filter keywords to overcome the large imbalance of nature data. This could introduce a bias to out-of-keyword but nature-related sentences.

\section*{Acknowledgements} 
This paper has received funding from the Swiss National Science Foundation (SNSF) under the project `How sustainable is sustainable finance? Impact evaluation and automated greenwashing detection' (Grant Agreement No. 100018\_207800).

\bibliography{main}

\newpage

\appendix

\section*{Appendix}

\section{A. Base Data for Labeling} \label{appendix:datacreation_steps_overview}
Table \ref{tab:baseLabelingFull} shows the full sentence characteristics of the base data for labeling.


\begin{table}[h]
\centering
\scalebox{0.9}{
\begin{tabular}{cccc}
\hline
\textbf{} & \textbf{AR} & \textbf{SR} & \textbf{EC} \\ \hline
\textbf{\# sentences} & 10.7 Mill. & 4.7 Mill. & 12.0 Mill. \\
\textbf{avg} & 29.2 & 22.7 & 18.2 \\
\textbf{std} & 21.5 & 11.8 & 12.4 \\
\textbf{min} & 1 & 1 & 1 \\
\textbf{25\%} & 18 & 15 & 9 \\
\textbf{50\%} & 25 & 21 & 16 \\
\textbf{75\%} & 26 & 25 & 25 \\
\textbf{max} & 229 & 1326 & 1016 \\ \hline
\end{tabular}}
\caption{Transposed View of Sentence Characteristics of Base Data for Labeling, collected from annual reports (AR), sustainability reports (SR), and earning calls transcripts (EC).}
\label{tab:baseLabelingFull}
\end{table}

\section{B. Sampling with Keywords} \label{appendix:keywords_sampeling}
During the subsequent phase of the data sampling for labeling, we strategically apply a set of chosen keywords, as illustrated in the subsequent list below. These keywords are selected for their broad scope, encompassing not only sentences directly pertaining to nature but also those that represent potential ambiguities, crucial for the discernment capabilities of the models in training. We sample 400,000 sentences of each data category of annual reports, sustainability reports, and earning call transcripts to have an equal representation of all data sources. Thus, we label 1.2 million sentences with the keywords. The keywords are sometimes reduced to their word stem, e.g., "environmental" is reduced to "environ".
\\

\begin{lstlisting}[frame=single, basicstyle=\ttfamily\tiny, breaklines=true, xleftmargin=0pt, numbers=none, label=code:prompt_1]
Water keywords: 'river', ' lake', ' aqua', 'h2o', ' rain', 'basin', 'reservoir', 'sanitation', 'drought', 'hydrat', 'dry', 'mineral', 'aquifer', 'glacier', 'glacial', 'fish stock', 'flood', 'precipitation', 'evapotr', 'groundwater', 'freshwater', 'water', 'ocean', 'marine', 'hurricane', 'coast', 'tsunami', 'ship', 'spill', 'vessel', 'cyclone', 'ENSO', 'El Nino', 'La Nina', 'storm', 'submerge', 'wind', 'sea', 'weather', 'fish'
Forest keywords: 'deforest', 'wood ', 'timber', 'ecosystem', 'raw material', ' tree', 'crop', 'cultivat', 'harvest', 'wild', 'flower', 'botanic', 'agriculture', ' farm', 'soy', 'leather', 'palm oil', 'paper', 'beef', 'pest', 'forestry', 'canopy', 'rotation', 'pulp', 'bark', 'fungi', 'forest'
Biodiversity keywords: 'animal', 'plant', 'bacteria', 'fungi', 'earth', 'extinct', 'biodivers', 'ecolog', 'insect', 'species', 'ecosystem', 'organism', 'forest', 'grassland', 'tundra', 'climat', 'tropic', 'rain forest', 'soil', 'cattle', 'cropland', 'farm', 'pollut', 'natur', 'mammal', 'bird', 'reptil', 'amphibian', 'environ', 'tree', 'rubber', 'miner', 'palm oil', 'soy ', 'hunt ', 'harvest', 'landscap', 'biolog', 'coral', 'habitat', 'biospher', 'biom', 'conservation', 'genet', 'national park', 'geograph', 'island', 'mountain', 'nativ', 'fauna', 'flora'
\end{lstlisting}

As Table \ref{tab:appkeysInall} suggests, the nature dimensions are minority classes representing only up to 10 \% of the data. However, even these numbers are highly inflated for two reasons. First, single keywords take a strong weight. As Figure \ref{fig:bioDivKeywords} suggests, keywords like "environ" take a majority weight in the set of biodiversity keywords. Second, the keywords themselves do not necessarily relate to nature. They are generally broader. To overcome this limitation, we use further data sampling steps to identify relevant text samples for labeling. 

\begin{table}[h!]
\centering
\scalebox{0.9}{
\begin{tabular}{cc}
\hline
\textbf{}             & \textbf{\begin{tabular}[c]{@{}c@{}}Appearance of keywords \\ in all sentences\end{tabular}} \\ \hline
\textbf{Water}        & 0.0836                                                                                      \\
\textbf{Forest}       & 0.0170                                                                                      \\
\textbf{Biodiversity} & 0.0991
\\ \hline
\end{tabular}}
\caption{Appearance of the keywords in all sentences}
\label{tab:appkeysInall}
\end{table}

\section{C. ChatGPT Pre-Labeling} \label{appendix:chatgpt_prompt}
\begin{figure}[htp]
        \centering
	\includegraphics[width=1\linewidth]{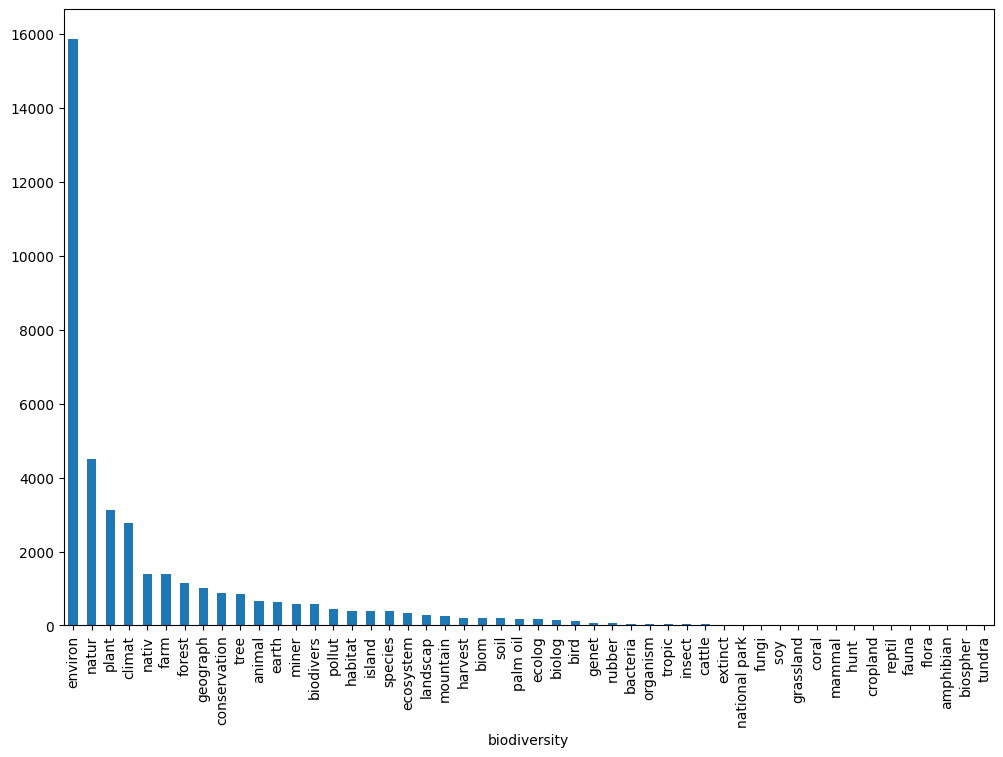}
 \caption{Keyword appearance for biodiversity keywords}
 \label{fig:bioDivKeywords}
\end{figure}

In the following, we display the prompt used to pre-label a subset of the keyword-filtered data. To reduce costs, we only label 5.000 text samples of each dimension. To effectively sample data and take into consideration the stark difference in the keyword appearance (see Figure \ref{fig:bioDivKeywords}), we build five buckets of keywords. We separate the keywords into the top 10\%, 10-20 \%, 20-40\%, 40-60\% and 60-100\%. For instance, "environ" is the most commonly appearing word in the biodiversity keywords. Thus, sentences containing this keyword would be assigned to bucket 1, the top 10\%. Then, we equally sample random sentences from each bucket to create a balance between the buckets. This aims to also sample sentences from minority keywords.

After pre-labeling the sentences with GPT-3.5, we get a score that indicates whether a sentence belongs to their respective class between water, forest, and biodiversity. We then differentiate between sentences that have a zero, low to middle (non-zero but lower than 75), and high (higher 75) scores to belong to the respective label. To construct the final label set for human expert annotators, we sample one-third of the data that belongs to zero, low to middle, and high to capture clear non-related samples, edge cases, and clearly related samples. Four human expert annotators label all these samples to confirm the intuition.
\\

\begin{lstlisting}[frame=single, basicstyle=\ttfamily\tiny, breaklines=true, xleftmargin=0pt, numbers=none, label=code:prompt_1]
"""
In the following you will be provided with a guideline enclosed in <> and a text enclosed in ||.
Your task is to label the text with the guideline. Read the text and assign a "Yes" if the text adheres to the guideline, "No" otherwise.

Please stricly follow the following answer format: answer with "Yes" or "No" and then provide a number between 0-100 of how sure you are (100 signaling very sure).

Provided guideline: <{guideline}>

Provided text: |{text}|
"""
\end{lstlisting}

\section{D. Labeling Guidelines} \label{appendix:labeling_guidelines}
We use the labeling guidelines displayed in Figure  \ref{tab:labelingGuidelines} for the prompts mentioned in Appendix C\ref{appendix:chatgpt_prompt} and for our expert human annotation process. The labeling guidelines are inspired by the reporting frameworks mentioned in the "Background" section. Specifically, we lean on the TNFD guidelines by defining the categories of water, forest, and biodiversity.

\begin{table}[htp]\scalebox{0.85}{
\begin{tabular}{|p{6cm}|p{3cm}|}                                                                                     \hline
\textbf{Water}                                                                                                                                                                                                                                                                                                                                                                                                                       & \textbf{No Water}                                                                                                                                 \\ \hline
The topic of water centers around water management, consumption, and pollution. It also concerns climate change effects on water, energy production with it, or projects integrating water issues. Also, sentences displaying direct impacts on water issues are considered like droughts.                                                                                                                                           & No effect on water issues. Simple generic environmental issues are not enough.                                                                    \\ \hline
\textbf{Forest}                                                                                                                                                                                                                                                                                                                                                                                                                      & \textbf{No Forest}                                                                                                                                \\ \hline
The forest topic comprises direct effects on timberland, trees, wood consumption, and related ecosystems. Also, the indirect effects of monocropping, and overharvesting leading to deforestation are considered. Additionally, working conditions, corruption, and bribery around the topic of forests are important.                                                                                                               & No Forest is just generic talk about nature without a direct or indirect link to forests. The sole address of environmental issues is not enough. \\ \hline
\textbf{Biodiversity}                                                                                                                                                                                                                                                                                                                                                                                                                & \textbf{No Biodiversity}                                                                                                                          \\ \hline
Biodiversity topics address the variety of living species on Earth, including plants, animals, bacteria, and fungi. A special focus lies on animal life in natural habitats and ecosystems. Categories of human interventions could be nature pollution, destruction, or restoration. Furthermore, special attention lies on hotspot areas of biodiversity loss. Hot spots may be coral reefs, rainforests, or other tropical areas. & No effect on biodiversity issues. Simple generic environmental issues are not enough.                                                             \\ \hline
\end{tabular}}
\caption{Labeling Guidelines}
\label{tab:labelingGuidelines}
\end{table}

\section{E. Inter-Annotator Agreement} \label{appendix:annotator_agreement}
For the labeling process, we employ human experts. Specifically, we select persons with a research background in sustainable finance. Three of the experts have PhDs on the topic and one is a PhD candidate at the time of labelling. Table \ref{tab:interannotator-agreement} shows an overview of the inter-annotator agreement between the four expert labelers. Overall, the agreement is very high and at a similar level for all three dimensions. This signals that the labeling task was fairly easy to perform for the labelers.

\begin{table}[h]\scalebox{0.9}{
\begin{tabular}{llrrr}
\hline
Dimension    & Fleiss' $\kappa$ & \multicolumn{1}{l}{2/4 Agree} & \multicolumn{1}{l}{3/4 Agree} & \multicolumn{1}{l}{4/4 Agree} \\ \hline
Water        & 0.822         & 0.023                         & 0.092                         & 0.884                         \\
Forest       & 0.799         & 0.023                         & 0.096                         & 0.881                         \\
Biodiversity & 0.834         & 0.023                         & 0.080                         & 0.897                         \\ \hline
\end{tabular}}
\caption{Inter-annotator agreement}
\label{tab:interannotator-agreement}
\end{table}

\section{F. Model Training and Results}\label{appendix:labeling_guidelines}  \label{appendix:modelresults}
Table \ref{tab:hyper} gives an overview of the base hyperparameter setup used to evaluate the baseline results in the five-fold cross-validation.

\begin{table}[h]
\centering
\begin{tabular}{cc}
\hline
\textbf{Hyperparameter}                                          & \textbf{Value} \\ \hline
Epochs                                                           & 10             \\
Batch Size                                                       & 8            \\
\begin{tabular}[c]{@{}c@{}}Gradient \\ Accumulation\end{tabular} & 2              \\
Warmup ratio                                                     & 0.1            \\
Learning rate                                                    & 5e-5           \\
Patience                                                         & 3              \\ \hline
\end{tabular}
\caption{Hyperparamters for training the models with five-fold cross-validation}
\label{tab:hyper}
\end{table} 
 
Table \ref{tab:modelresults} shows an overview of the models' F1-score, accuracy, recall, and precision. It becomes apparent that the further pre-trained models on the climate- and environmental domain slightly outperform their (larger) counterparts.

\begin{table*}[h]
\centering

\begin{tabular}{cccccc}
\hline
\textbf{Domain}                        & \textbf{Model}             & \textbf{F1-Score (std.)} & \textbf{Accuracy (std.)} & \textbf{Precision (std.)} & \textbf{Recall (std.)}   \\ \hline
\multirow{4}{*}{{Water}}        & {EnvironmentalBERT} & 0.9447 (0.0137)          & 0.9591 (0.0104)          & 0.9413 (0.0166)           & 0.9486 (0.0134)          \\
                                       & {ClimateBERT}       & \textbf{0.9510 (0.0113)}  & \textbf{0.9636 (0.0088)} & 0.9470 (0.0165)            & \textbf{0.9555 (0.0074)} \\
                                       & {RoBERTa}           & 0.9455 (0.0086)          & 0.9600 (0.0065)            & \textbf{0.9452 (0.0140)}   & 0.9478 (0.0110)           \\
                                       & {DistilRoBERTa}     & 0.9498 (0.0116)          & 0.9627 (0.0087)          & 0.9445 (0.0149)           & 0.9552 (0.0130)           \\ \hline
\multirow{4}{*}{{Forest}}       & {EnvironmentalBERT} & \textbf{0.9537 (0.0092)} & \textbf{0.9700 (0.0061)}   & \textbf{0.9509 (0.0111)}  & 0.9567 (0.0103)          \\
                                       & {ClimateBERT}       & 0.9534 (0.0046)          & 0.9695 (0.0034)          & 0.9479 (0.0132)           & \textbf{0.9598 (0.0080)}  \\
                                       & {RoBERTa}           & 0.9478 (0.0048)          & 0.9664 (0.0025)          & 0.9470 (0.0076)            & 0.9494 (0.0141)          \\
                                       & {DistilRoBERTa}     & 0.9529 (0.0065)          & 0.9695 (0.0041)          & 0.9509 (0.0099)           & 0.9556 (0.0132)          \\ \hline
\multirow{4}{*}{{Biodiversity}} & {EnvironmentalBERT} & \textbf{0.9276 (0.0191)} & \textbf{0.9582 (0.0105)} & \textbf{0.9285 (0.0176)}  & \textbf{0.9278 (0.0267)} \\
                                       & {ClimateBERT}       & 0.9249 (0.0134)          & 0.9564 (0.0074)          & 0.9242 (0.0153)           & 0.9267 (0.0220)           \\
                                       & {RoBERTa}           & 0.9246 (0.0154)          & 0.9564 (0.0083)          & 0.9243 (0.0116)           & 0.9256 (0.0236)          \\
                                       & {DistilRoBERTa}     & 0.9229 (0.0203)          & 0.9545 (0.0118)          & 0.9158 (0.0202)           & 0.9307 (0.0212)          \\ \hline
\multirow{4}{*}{{Nature}}       & {EnvironmentalBERT} & \textbf{0.9419 (0.0081)} & \textbf{0.9423 (0.0080)}  & \textbf{0.9444 (0.0070)}   & \textbf{0.9409 (0.0085)} \\
                                       & {ClimateBERT}       & 0.9350 (0.0046)           & 0.9355 (0.0044)          & 0.9380 (0.0035)            & 0.9339 (0.0051)          \\
                                       & {RoBERTa}           & 0.9397 (0.0026)          & 0.9400 (0.0026)            & 0.9418 (0.0028)           & 0.9386 (0.0027)          \\
                                       & {DistilRoBERTa}     & 0.9355 (0.0072)          & 0.9359 (0.0071)          & 0.9381 (0.0066)           & 0.9345 (0.0075)          \\ \hline
\end{tabular}
\caption{F1-score, Accuracy, Recall, and Precision for base hyperparameters setup for training the models with five-fold cross-validation}
\label{tab:modelresults}
\end{table*}

\section{G. Hyperparameter Grid} \label{appendix:hypergrid}
To check whether the model performance stays consistent for different hyperparameters, we vary the learning rate and batch size in the fine-tuning process. As Table \ref{tab:hyperpargrid} shows, all models perform on a very similar level irrespective of the hyperparameter setup. However, the pre-trained models ClimateBERT and EnvironmentalBERT consistently outperform their counterparts.

\begin{table*}[h]
\centering
\begin{tabular}{lcccccc}
\hline
\multicolumn{1}{c}{\textbf{Dimension}}       & \multicolumn{2}{c}{\textbf{Hyperparameters}}  & \multicolumn{4}{c}{\textbf{Model F1-score (std.)}}                                                        \\ \hline
                                            & {learning rate} & {batch size} & {EnvironmentalBERT} & {ClimateBERT}     & {RoBERTa}         & {DistilRoBERTa} \\ \hline
\multicolumn{1}{c}{\multirow{4}{*}{Water}}  & 3e-05                  & 8                   & 0.9465 (0.0145)            & \textbf{0.9539 (0.0118)} & 0.9437 (0.0070)           & 0.9478 (0.0110)         \\
\multicolumn{1}{c}{}                        & 3e-05                  & 16                  & 0.9459 (0.0120)             & \textbf{0.9488 (0.0136)} & 0.9470 (0.0115)           & 0.9481 (0.0122)        \\
\multicolumn{1}{c}{}                        & 5e-05                  & 8                   & 0.9447 (0.0137)            & \textbf{0.9510 (0.0113)}  & 0.9455 (0.0086)          & 0.9498 (0.0116)        \\
\multicolumn{1}{c}{}                        & 5e-05                  & 16                  & 0.9472 (0.0098)            & \textbf{0.9510 (0.0113)}  & 0.9441 (0.0124)          & 0.9482 (0.0122)        \\ \hline
\multicolumn{1}{c}{\multirow{4}{*}{Forest}} & 3e-05                  & 8                   & \textbf{0.9532 (0.0109)}   & 0.9512 (0.0153)          & 0.9531 (0.0103)          & 0.9499 (0.0151)        \\
\multicolumn{1}{c}{}                        & 3e-05                  & 16                  & \textbf{0.9534  (0.0138)}  & 0.9500 (0.0122)            & 0.9512 (0.0116)          & 0.9529 (0.0065)        \\
\multicolumn{1}{c}{}                        & 5e-05                  & 8                   & \textbf{0.9537 (0.0092)}   & 0.9534 (0.0046)          & 0.9478 (0.0048)          & 0.9529 (0.0065)        \\
\multicolumn{1}{c}{}                        & 5e-05                  & 16                  & \textbf{0.9587 (0.0132)}   & 0.9497 (0.0123)          & 0.9499 (0.0151)          & 0.9521 (0.0124)        \\ \hline
\multicolumn{1}{c}{\multirow{4}{*}{Biodiversity}}               & 3e-05                  & 8                   & \textbf{0.9285 (0.0159)}   & 0.9243 (0.0255)          & 0.9292 (0.0161)          & 0.9197 (0.0125)        \\
                                            & 3e-05                  & 16                  & \textbf{0.9246 (0.0143)}   & 0.9245 (0.0188)          & 0.9230 (0.0145)           & 0.9203 (0.0132)        \\
                                            & 5e-05                  & 8                   & \textbf{0.9276 (0.0191)}   & 0.9249 (0.0134)          & 0.9246 (0.0154)          & 0.9229 (0.0203)        \\
                                            & 5e-05                  & 16                  & 0.9239 (0.0141)            & \textbf{0.9263 (0.0211)} & 0.9167 (0.0135)          & 0.9207 (0.0080)         \\ \hline
\multicolumn{1}{c}{\multirow{4}{*}{Nature}}                     & 3e-05                  & 8                   & \textbf{0.9382 (0.0090)}    & 0.9332 (0.0063)          & 0.9372 (0.0078)          & 0.9365 (0.0072)        \\
                                            & 3e-05                  & 16                  & \textbf{0.9382 (0.0131)}   & 0.9332 (0.0063)          & 0.9374 (0.0072)          & 0.9345 (0.0104)        \\
                                            & 5e-05                  & 8                   & \textbf{0.9419 (0.0081)}   & 0.935 (0.0046)           & 0.9397 (0.0026)          & 0.9355 (0.0072)        \\
                                            & 5e-05                  & 16                  & 0.9369 (0.0074)            & 0.9355 (0.0051)           & \textbf{0.9392 (0.0057)} & 0.9308 (0.0140)         \\ \hline
\end{tabular}
\caption{F1-score for a variation of hyperparameters for training the models with five-fold cross-validation}
\label{tab:hyperpargrid}
\end{table*}

\section{H. Biodiversity Keyword Approach} \label{appendix:biodivkeywords}
Nature has been sparsely explored in prior literature. One contribution to biodiversity has been developed by \citet{giglio2023}. They use a keyword approach to classify sentences belonging to biodiversity in annual reports. Specifically, a two-layered approach is used to detect biodiversity text samples. A sentence must contain both one \textit{specific} and one \textit{additional} keyword. See the listing below for the keywords:
\\
\begin{lstlisting}[frame=single, basicstyle=\ttfamily\tiny, breaklines=true, xleftmargin=0pt, numbers=none, label=code:prompt_1]
Specific keywords: 
"biodiversity", "ecosystem", "ecology", "ecological", "habitat", "species", "forest", "deforestation", "fauna", "flora", "marine", "tropical", "freshwater", "wetland", "wildlife", "coral", "aquatic", "desertification", "carbon sink", "ecosphere", "biosphere"

Additional keywords: 
Ecosystem: "climate", "coast", "forest", "micro", "natur", "public health", "sustaina", "water"
Marine: "marine biodiversity", "marine ecosystem", "marine environment", "marine life", "marine species"
Tropical: "tropical biodiversity", "tropical ecosystem", "tropical environment", "tropical forest", "tropical species"
Species: "aquatic", "biodiversity", "bird", "endanger", "environment", "fish", "habitat", "invasive", "list", "marine", "protect", "threat", "ESA", "EPA"
\end{lstlisting}

We label our expert-annotated dataset with these keywords and calculate the previously employed performance metrics. As Table \ref{tab:labelingGuidelines} shows, we come to an interesting yet sobering conclusion: the ML-based approach outperforms the keyword approach by a significant margin. There are two plausible reasons for this. First, in line with prior literature, keyword approaches generally underperform BERT-based approaches \citep{varini2020}. This lies in their inability to assess context. Particularly in a complex topic like biodiversity, this might play a role. Second, the scope of our biodiversity label is very conservative (see Table \ref{tab:labelingGuidelines}). We only take into account direct biodiversity topics or immediate indirect effects such as ecosystem pollution. Keywords are by their nature broader. The result is that keyword-based approaches may contain a large amount of noise, despite still proving useful for the construction of hedging portfolios, as in \citet{giglio2023}.

\begin{figure*}[h]
\centering
\includegraphics[width=0.5\linewidth]{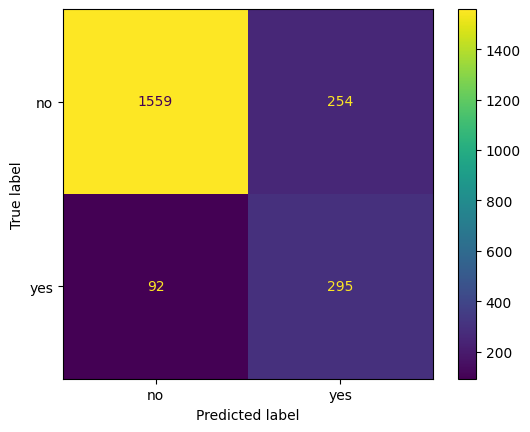}
\caption{Confusion matrix for classifying biodiversity sentences with the biodiversity keyword approach}
\label{fig:confmat_biodidv}
\end{figure*}

Evidence for this result can be found in the misclassifications. As Figure \ref{fig:confmat_biodidv} shows, the keyword approach suffers from classifying non-biodiversity sentences with the true label (false positives). The following sentence from our dataset exemplifies the underlying reason: \enquote{Together, our \textbf{forests} and products play an important role in mitigating \textbf{climate} change by limiting the amount of carbon dioxide that is released into the atmosphere each year}. According to the keyword approach, \enquote{forest} and \enquote{climate} in one sentence signal biodiversity. In our definition, this sentence disqualifies as a biodiversity sentence because it is too broad to signal that the company has a direct impact on, or is impacted by, biodiversity. While there are interconnections between climate change, forest, and biodiversity, these sentences are too vague and do not entail direct effects on the variety of plant or animal species. This is why we also check if the keyword approach rather captures a general concept of nature. However, as Table \ref{tab:keywordApproach} shows, this is not the case.

We are cautious to discredit the keyword approach entirely. However, nature is a complex system that is difficult to predict \footnote{See for instance: Diamond (2011): How societies choose to fail or
succeed: revised edition.}, and yet is important. We argue this warrants a careful approach when classifying companies' relationship with nature. This is our main contribution. Further, classification becomes important when companies can utilize vague, uninformative language when discussing nature risks\footnote{See for instance: ISSB (2022): Agenda item 1a.}, essentially adding so much noise as to become equivalent to non-disclosure. 

A NLP-based approach may also not be a panacea. Other approaches, such as the use of satellites, offer novel additional ways to assess the relationship between companies and nature. This warrants future attention.

\begin{table*}[h]
\centering
\begin{tabular}{ccccc}
\hline
\textbf{Keywords}                                               & \textbf{F1-score} & \textbf{Accuracy} & \textbf{Precision} & \textbf{Recall} \\ \hline
\begin{tabular}[c]{@{}c@{}}on biodiversity label\end{tabular} & 0.6303            & 0.8427            & 0.7623             & 0.5373          \\
\begin{tabular}[c]{@{}c@{}}on nature label\end{tabular}       & 0.6100            & 0.6978            & 0.4498             & 0.9472          \\ \hline
\end{tabular}
\caption{Results of applying the keyword approach on the entirety of the dataset}
\label{tab:keywordApproach}
\end{table*}

\section{I. Nature Communication in Industries and Countries} \label{appendix:naturemention}
Figure \ref{fig:natureIndustries} shows an overview of the top 20 industries that discuss nature, based on the Fama-French 49 industry classifications. Somewhat intuitively, the industries that most discuss nature are the Agriculture and Utility industries. Resource-intensive industries such as Coal and Construction also discuss nature comparatively frequently. These results solidify the validity of the developed models.

\begin{figure*}[h]
\centering
\includegraphics[width=1\linewidth]{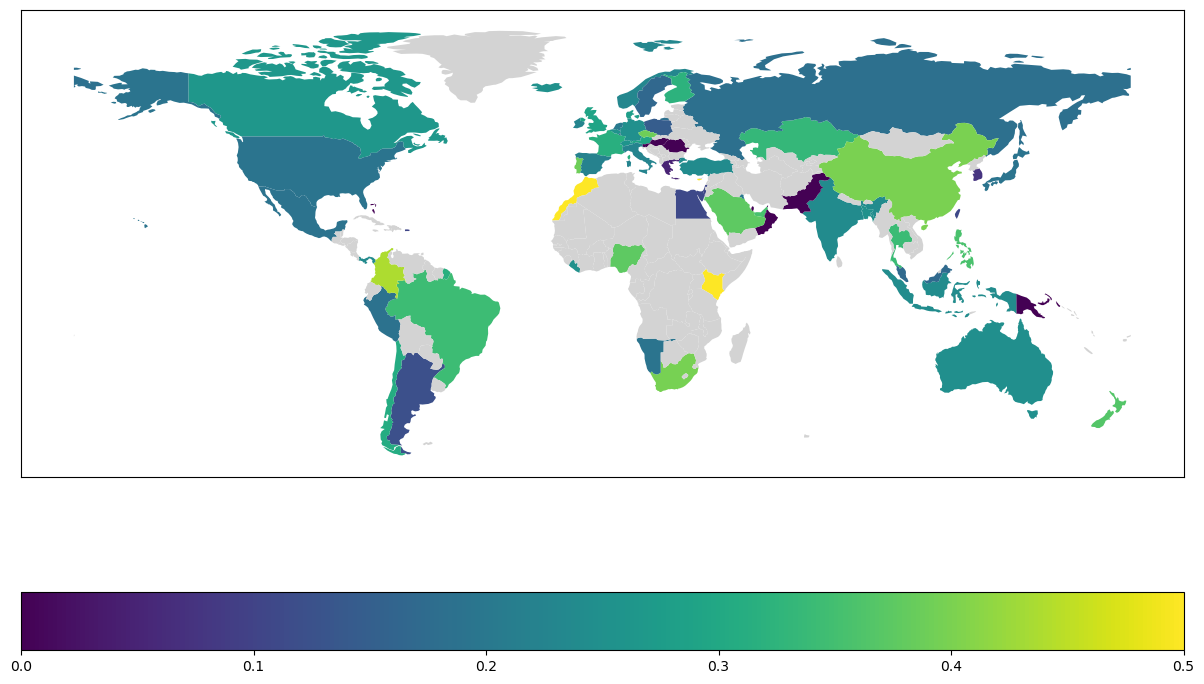}
\caption{Proportion of earnings conference calls that mention nature-related topics in at least one sentence in each country}
\label{fig:natureMentioning}
\end{figure*}

Figure \ref{fig:natureMentioning} shows the proportion of earning calls that mention nature-related topics at least once. As the figure suggests, nature is mostly discussed in ecosystem hotspots near the equator such as Brazil and Indonesia which heavily rely on forestry or agriculturally dominated countries in the global south.

We refrain from further interpretations as this case study only presents the first advances in analyzing nature communication of companies at a broader scale.

\end{document}